\documentclass[11pt]{article}
\newif\ifarxiv
\arxivtrue
\newif\ifcamera
\cameratrue
\newif\ifnldp
\nldpfalse

\expandafter\ifx\csname ifcamera\endcsname\relax
  \expandafter\newif\csname ifcamera\endcsname
\fi

\ifnldp
\else
  \ifarxiv
    \usepackage{acl}
  \else
    \usepackage[review]{acl}
    \usepackage{zref-savepos}
    \makeatletter
    \newdimen\@LN@midpt
    \AtBeginDocument{%
      \@LN@midpt=\dimexpr\hoffset + 1in + \oddsidemargin + \textwidth/2\relax}
    \def\makeLineNumberOuter{%
      \edef\@LN@key{LN\the\c@linenumber}%
      \zsavepos{\@LN@key}%
      \ifdim\zposx{\@LN@key}sp<\@LN@midpt
        \makeLineNumberLeft
      \else
        \makeLineNumberRight
      \fi}

    \AtBeginDocument{}
    \makeatother
  \fi
\fi

\usepackage{xurl}
\ifnldp
\else
  \usepackage{times}
  \usepackage{inconsolata}
\fi
\usepackage{latexsym}
\usepackage[T1]{fontenc}
\usepackage[utf8]{inputenc}
\usepackage{graphicx}
\usepackage{booktabs}
\usepackage{tabularx}
\usepackage{ragged2e}
\usepackage{array}
\usepackage{adjustbox}
\usepackage{threeparttable}
\usepackage{enumitem}
\usepackage{amsmath}
\newcommand{\fullblock}{\rule{0.8ex}{0.8ex}}
\DeclareUnicodeCharacter{2588}{\fullblock}


\title{FrenchNews-7: Benchmarking Cross-Publisher French News Editorial Desk Classification}

\ifnldp
  \author{Amr Sobhy}
  \institute{Le French News Lab \\ \email{amr@frenchnewslab.org}}
\else
  \ifarxiv
    \author{Amr Sobhy \\ Le French News Lab \\ \texttt{amr@frenchnewslab.org}}
  \else
    \author{Anonymous ARR Submission}
  \fi
\fi

\begin{document}
\maketitle

\begin{abstract}
We present FrenchNews-7, a cross-publisher France-based French-language news editorial desk classification benchmark combining a large multi-outlet corpus, a URL-derived seven-class taxonomy, and a fine-tuned CamemBERT classifier. Labels are assigned via a hybrid pipeline combining publisher URL slugs with LLM annotation for structurally ambiguous cases, audited through an inter-rater study (2 humans + 2 LLMs; pairwise $\kappa \geq 0.766$, human--human $\kappa = 0.806$). We evaluate lexical, multilingual, and French-specific trained classifiers under both in-distribution and held-out-publisher settings, with additional comparison against zero-shot LLM baselines (GPT-OSS-120B, Mistral Small~3.2, Llama-3.3-70B) on the held-out pool. The strongest model, CamemBERT-base on full article text, outperforms headline-only input, generalizes to unseen outlets, and exceeds all three zero-shot LLM baselines on overall recall (0.799), with the gap concentrated in the ambiguous editorial-boundary categories \emph{Économie} and \emph{Société}. Cross-publisher evaluation reveals uneven boundary stability: \emph{Sport}, \emph{Culture \& Loisirs}, and \emph{International} transfer cleanly, while \emph{Économie} (recall~=~0.517) is statistically indistinguishable from blinded human agreement on the same articles (0.55, $n = 40$), and \emph{Société} (precision~=~0.577) absorbs boundary ambiguity, both suggesting editorial conventions rather than recoverable classifier headroom.
\ifarxiv
The fine-tuned CamemBERT-base model is released on Hugging Face (\url{https://huggingface.co/LeFrenchNewsLab/camembert-base-frenchnews7}) and the labeled manifest on (\url{https://huggingface.co/datasets/LeFrenchNewsLab/frenchnews-7}), alongside reference collection scripts to assist re-fetching and a reliability-tier guidance table.
\else
The fine-tuned CamemBERT-base model, labeled manifest, reference collection scripts, and a reliability-tier guidance table are available at [anonymous link].
\fi
\end{abstract}

\ifarxiv\ifcamera\else
\fi\fi

\section{Introduction}

Computational comparative media research on French news (agenda-setting studies, media diversity audits, longitudinal topic coverage analysis) requires a classifier that works reliably across publishers. Existing French-language news corpora are either single-outlet \citep{scialom2020,wastl2025} or too narrow in scope to generalize \citep{icard2024}, and to our knowledge, no prior French benchmark targets this cross-publisher setting directly. FrenchNews-7 fills this gap, providing editorial desk classification infrastructure grounded in the French news ecosystem: a labeled corpus of 87,637 articles from 13 France-based outlets, paired with multi-model benchmarks and a held-out cross-publisher evaluation.

The label is a harmonized editorial-desk category: inferred from the publisher's URL slug when available (72.2\%), and content-labeled by an LLM into the same taxonomy when the URL carries no desk signal (27.8\%). The label \emph{space} overlaps heavily with conventional news topics: a crosswalk to IPTC Media Topics (Appendix~\ref{app:iptc_crosswalk}) maps 10 of 17 top-level IPTC categories onto FrenchNews-7 classes. The difference is in the label-generating process: the target variable is the publisher's routing decision, recorded at collection, not an annotator's judgement of topic. \emph{International} has no IPTC equivalent because it encodes geographic routing rather than subject matter. The benchmark measures how reliably an automated classifier can recover these editorial routing decisions from article text alone, a task with direct applications in computational journalism, where downstream analyses depend on scaling editorial-section assignment across outlets. Building a cross-publisher benchmark requires two practical decisions: what label space to use, and how to obtain labels at scale. For the label space, we use URL-structure analysis of the 13 outlets in this corpus to identify seven categories that are (a) present in at least 10 of 13 publishers and (b) each account for at least 3.5\% of total output. The resulting classes are \emph{Société} (society and domestic affairs), \emph{Culture \& Loisirs} (culture and leisure), \emph{International}, \emph{Politique} (domestic politics), \emph{Sport}, \emph{Économie} (economy), and \emph{Sciences \& Technologies}. For labeling, a hybrid pipeline (\S\ref{sec:annotation_pipeline}) combines deterministic URL-slug rules with LLM annotation for ambiguous cases.

FrenchNews-7 contributes an editorial desk classification corpus and a CamemBERT classifier evaluated under both in-distribution and held-out-publisher conditions (\S\ref{sec:main_results}, \S\ref{sec:crosspub_gen}).

\section{Related Work}

\paragraph{News topic classification.} Early TF-IDF + SVM approaches \citep{joachims1998} established strong benchmark settings for text categorization; \citet{lewis2004} formalized RCV1 as the large-scale standard benchmark for the field. Later neural work expanded to large-scale datasets such as AG News \citep{zhang2015}. BERT-family encoders \citep{devlin2019,liu2019,he2021} then achieved substantial performance gains, with attention shifting to harder settings: hierarchical classification, cross-lingual transfer, and language-specific news classification. Our challenge is different: cross-publisher generalization without manual annotation at scale.

\paragraph{French NLP.} CamemBERT \citep{martin2020} is a RoBERTa-based encoder pre-trained on \textasciitilde138GB of French text, achieving state-of-the-art results across French benchmarks. CamemBERTav2 \citep{antoun2024} updates the architecture (DeBERTaV3) and training data (275B tokens), extending the sequence limit to 1,024 tokens. FlauBERT \citep{le2020} provided the FLUE benchmark suite. The closest prior French news classifier we identified is \texttt{lincoln/flaubert-mlsum},\footnote{Available at huggingface.co/lincoln/flaubert-mlsum (Hugging Face Model Hub).} a FlauBERT model fine-tuned on MLSUM \citep{scialom2020} whose 10-class taxonomy is URL-derived from \emph{Le Monde} slugs; as \emph{Le Monde} constitutes 37.8\% of our corpus, direct comparison introduces contamination. \citet{pelloin2024} classify French broadcast transcripts into 18 IPTC-inspired categories using a teacher-student approach; their work is the closest methodological precedent but targets audio content from 21 broadcast channels. On the written-press side, \citet{escouflaire2024} distinguish opinion from factual news in Belgian and Canadian French press using CamemBERT vs.\ feature-based classifiers (a related but orthogonal genre classification problem): binary opinion vs.\ factual-news detection rather than cross-publisher editorial desk routing.

\paragraph{LLM-based annotation.} \citet{gilardi2023} showed zero-shot ChatGPT outperforms crowd workers across multiple annotation tasks while costing around 30$\times$ less per label. \citet{tornberg2023} showed GPT-4 matches or exceeds expert annotators on political text. \citet{pangakisw2024acl} demonstrate GPT-4-labeled fine-tuned classifiers match human-labeled baselines across 14 tasks. Our Bucket B applies this paradigm only to structurally ambiguous articles (27.8\%), while the majority of labels are deterministic. \citet{lu2025} warn of non-random error propagation in LLM-labeled models; our hybrid design mitigates this by anchoring 72.2\% of labels in publisher editorial decisions.

\paragraph{Cross-publisher harmonization.} Individual publishers use proprietary taxonomies. The closest work we identified is \citet{kuzman2025}: EMMediaTopic annotates 21,000 texts in four non-French languages into 17 IPTC categories using GPT-4o, then trains XLM-RoBERTa-large (macro-F1 $= 0.746$). The \texttt{lincoln/flaubert-mlsum} model card (ibid.) provides the single-publisher precedent for URL slug--derived taxonomy; FrenchNews-7 extends this signal to 13 publishers and introduces cross-outlet harmonization as a distinct challenge. We did not identify prior work that combines URL-derived taxonomy, multi-outlet French data, hybrid annotation, and held-out cross-publisher evaluation in a single benchmark.

\paragraph{Taxonomy comparability.} Rather than treat taxonomy misalignment as a barrier, we quantify it by probing the EMMediaTopic classifier \citep{kuzman2025} on the FrenchNews-7 test split without retraining, a worst-case probe: French text, an unseen label space. Structural taxonomy misalignment accounts for less than half the observed performance gap relative to our CamemBERT model; the dominant factor is cross-lingual domain shift (see Appendix~\ref{app:iptc_experiment} for full results). The largest taxonomy collision is instructive: ``conflict, war and peace'' (an IPTC category with no FrenchNews-7 equivalent) is the third most frequent prediction (9.3\% of test articles), absorbing French geopolitical content that FrenchNews-7 assigns to \emph{International} or \emph{Politique}. The performance gap reflects structural taxonomy mismatch rather than representational limitations, a distinction that model retraining alone cannot address.

\section{Dataset Construction}

\subsection{Taxonomy Design}

To construct a label space suitable for cross-publisher classification, we analyze the URL structures of the 13 publishers in this corpus rather than adopting an external ontology. Each publisher encodes its editorial sections directly in article URL path segments (e.g., \texttt{lemonde.fr/politique/}, \texttt{lexpress.fr/sport/}). We extract the first path segment from all 87,637 articles and group them into semantic families by editorial intent, yielding 13 candidate families ranked by volume and cross-publisher coverage.

Table~\ref{tab:taxonomy_derivation} shows the result. We apply two inclusion criteria: (a) presence in $\geq 10$ of 13 publishers and (b) corpus share $\geq 3.5\%$; both must hold. The 3.5\% threshold falls in the 1.3pp gap between the seventh-ranked family (\emph{Sciences \& Technologies}: 3.5\%, 10/13 publishers) and the eighth (\emph{Environnement}: 2.2\%, 10/13), the largest drop between adjacent candidates in the upper ranking. Below the seventh family, candidates lack either volume (\emph{Santé} 1.2\%, \emph{Médias} 1.0\%) or cross-publisher consensus (\emph{Régional}: 5 publishers, \emph{Opinion}: 6). These criteria were applied at taxonomy-design time, before any classifier was trained, and the threshold neighbourhood is fully determined by this ranking: a 2\% cut would add only \emph{Environnement} and a 4\% cut would drop only \emph{Sciences \& Technologies}. We therefore do not retrain reduced or expanded label-space variants.

\begin{table}[t]
\centering
\small
\setlength{\tabcolsep}{4pt}
\renewcommand{\arraystretch}{1.10}
\begin{tabular}{@{}lrrc@{}}
\toprule
\textbf{Family} & \textbf{Articles} & \textbf{\% corpus} & \textbf{Publishers} \\
\midrule
International & 16,090 & 18.0\% & 13/13 \\
Culture/Loisirs & 15,637 & 17.5\% & 13/13 \\
Politique & 11,513 & 12.9\% & 13/13 \\
Société & 11,234 & 12.6\% & 13/13 \\
Économie & 8,233 & 9.2\% & 13/13 \\
Sport & 7,178 & 8.0\% & 13/13 \\
Sciences \& Tech. & 3,144 & 3.5\% & 10/13 \\
\midrule
Environnement & 1,976 & 2.2\% & 10/13 \\
Santé & 1,030 & 1.2\% & 9/13 \\
Médias & 903 & 1.0\% & 11/13 \\
Opinion/Édito & 873 & 1.0\% & 6/13 \\
Régional & 1,366 & 1.5\% & 5/13 \\
Format/Misc & 5,570 & 6.2\% & 11/13 \\
\bottomrule
\end{tabular}
\caption{Candidate URL families ranked by volume. The rule marks the seven-family break; lower families are absorbed into semantic parents or deferred to Bucket B. The 2,890 articles with absent or unparseable URL paths are labeled via Bucket B.}
\label{tab:taxonomy_derivation}
\end{table}

Absorption decisions for families below the break: \emph{Santé} and \emph{Environnement} fold into \emph{Société} (domestic social affairs encompasses health and local environmental content); \emph{Médias} folds into \emph{Culture \& Loisirs} (media criticism and television coverage); \emph{Opinion} and \emph{Régional} slugs carry no unambiguous thematic signal and are deferred to Bucket B LLM labeling. Any discrete taxonomy involves pragmatic boundary decisions; we report ours transparently so that future work can extend the label space (e.g., promoting \emph{Environnement} or \emph{Santé} to standalone classes) without invalidating the existing seven-class annotations.

The seven categories are: \emph{Société} (society: domestic social affairs, health, education, crime), \emph{Culture \& Loisirs} (culture and leisure: arts, cinema, books, leisure), \emph{International} (foreign affairs, geopolitics), \emph{Politique} (domestic politics, institutions), \emph{Sport}, \emph{Économie} (economy: business, finance, markets), and \emph{Sciences \& Technologies} (science and technology: research, technology, digital). \emph{Environnement} (environment) is not a standalone class; environmental content is distributed across \emph{Société} and \emph{Sciences \& Technologies}. Appendix~\ref{app:iptc_crosswalk} provides a mapping of these seven categories to the top-level IPTC Media Topics standard, documenting the convergence between a bottom-up empirical derivation and an independently developed international standard.

\subsection{Data Collection}

We collected 87,769 articles from 13 France-based French-language media outlets. These outlets were selected based on public archive availability and the need to construct a structurally representative cross-section of the French media landscape, deliberately spanning national daily and weekly press, political magazines, regional press, digital-native outlets, and broadcast news, covering primarily 2018--2026, with a sparse archive tail extending to 2005 (4\% of articles pre-date 2018). After deduplication by SHA-256 hash of article text (removing 132 duplicate records), the final corpus contains 87,637 articles; the non-empty headline and body filter removed no further records. Each article record stores the headline, full body text, publication date, source URL, and the publisher's original URL slug.

Table~\ref{tab:publishers} lists all 13 outlets. \emph{Le Monde} dominates at 37.8\%; the remaining twelve contribute 2.1--10.4\% each. Bucket A coverage ranges from 100\% (JDD) to 34.3\% (Slate.fr).

\begin{table}[t]
\centering
\footnotesize
\setlength{\tabcolsep}{3pt}
\renewcommand{\arraystretch}{1.06}
\begin{tabular}{@{}lrrrr@{}}
\toprule
\textbf{Publisher} & \textbf{Articles} & \textbf{\%} & \textbf{Type} & \textbf{Bucket A} \\
\midrule
Le Monde & 33,090 & 37.8 & NP & 63.2 \\
L'Express & 9,093 & 10.4 & NP & 82.6 \\
L'Humanité & 7,975 & 9.1 & NP & 68.6 \\
Le Point & 7,877 & 9.0 & NP & 87.6 \\
Le Figaro & 7,270 & 8.3 & NP & 63.0 \\
JDD & 7,269 & 8.3 & NP & 100.0 \\
La Croix & 1,910 & 2.2 & NP & 52.7 \\
Le HuffPost & 2,473 & 2.8 & DN & 84.0 \\
TF1 INFO & 2,044 & 2.3 & BR & 97.4 \\
Le Parisien & 2,375 & 2.7 & NP & 82.9 \\
Slate.fr & 2,105 & 2.4 & DN & 34.3 \\
20 Minutes & 2,291 & 2.6 & DN & 94.9 \\
Ouest-France & 1,865 & 2.1 & RP & 39.0 \\
\midrule
\textbf{Total} & \textbf{87,637}$^*$ & \textbf{100} & \textbf{13} & \\
\bottomrule
\end{tabular}
\caption{Corpus composition by publisher. Type: NP = national press, RP = regional press, DN = digital-native, BR = broadcast web. Bucket A = deterministic slug-label share. $^*$Post-deduplication; 132 duplicates removed.}
\label{tab:publishers}
\end{table}

The ideological diversity of the outlet set (spanning left [\emph{L'Humanité}], centre-left [\emph{Le Monde}, \emph{HuffPost}], centre [\emph{JDD}, \emph{TF1 INFO}], centre-right [\emph{L'Express}, \emph{Le Point}], and right [\emph{Le Figaro}]) ensures that category boundaries are learned from politically heterogeneous coverage, mitigating the risk that the taxonomy learns politically-inflected rather than topically-grounded distinctions.

\subsection{Two-Bucket Annotation Pipeline}
\label{sec:annotation_pipeline}

Annotation follows a two-bucket pipeline that prioritizes deterministic label assignment over model judgment wherever possible.

\paragraph{Bucket A: Slug-based labeling (63,302 articles, 72.2\%).}
For articles whose URL path slug maps unambiguously to a taxonomy category, the label is assigned by a deterministic lookup table of 74 rules. Rules are organized by category: 20 slugs map to \emph{Culture \& Loisirs} (e.g., \texttt{cinema}, \texttt{musique}, \texttt{livres}), 10 to \emph{International} (e.g., \texttt{monde}, \texttt{europe}, \texttt{afrique}), 7 to \emph{Sport} (e.g., \texttt{sport}, \texttt{rugby}, \texttt{jo-paris-2024}), 8 to \emph{Sciences \& Technologies}, 2 to \emph{Politique}, 6 to \emph{Économie}, and 21 to \emph{Société}.
Publisher slug-coverage rates (Table~\ref{tab:publishers}) range from 100\% (JDD) to 34.3\% (Slate.fr).

\paragraph{Bucket B: LLM labeling (24,335 articles, 27.8\%).}
Articles whose URL slug carries no unambiguous category signal fall into Bucket B. Three slug types trigger LLM labeling: (1) cross-topic editorial desks (\texttt{/idees}, \texttt{/debats}, \texttt{/les-decodeurs}, \texttt{/m-le-mag}); (2) format or catch-all slugs (\texttt{/story}, \texttt{/flash-actu}, \texttt{/informations}); (3) regional or absent slugs (\texttt{/pays-de-la-loire}, broken paths). For these articles, GPT-OSS-120B is presented with a prompt consisting of the seven category definitions (one sentence per class), the article headline, and the full body text, and required to output exactly one category name with temperature $= 0$ (deterministic decoding). Direct annotation follows \citet{ding2023}, who find that prompt-guided labeling is well-suited to tasks with small, well-defined label spaces, precisely the structure of our seven-class taxonomy. We use GPT-OSS-120B because it is open-weight and freely accessible, and large enough (117B parameters, 5.1B active per token) to be adequate for this labelling task rather than a small low-cost model; the Bucket~B pass is therefore re-runnable by others from the released prompt and labels, with no dependence on a hosted model. A bucket ablation study (\S\ref{sec:annotation_quality}) confirms that these labels do not affect reported results.

The combined label distribution is shown in Table~\ref{tab:dataset_dist}. Bucket B contributes meaningfully to all categories, with the largest relative contribution to \emph{Sciences \& Technologies} (2,215 Bucket B articles vs.\ 1,982 Bucket A, reflecting that science slugs are less consistently represented across publishers).

\begin{table}[t]
\centering
\scriptsize
\setlength{\tabcolsep}{4pt}
\renewcommand{\arraystretch}{1.12}
\begin{tabular}{@{}lrrrrr@{}}
\toprule
\textbf{Category} & \textbf{Bucket A} & \textbf{Bucket B} & \textbf{Total} & \textbf{\%} \\
\midrule
Société & 12,666 & 6,658 & 19,324 & 22.1\% \\
International & 15,510 & 3,132 & 18,642 & 21.3\% \\
Culture \& Loisirs & 13,305 & 5,045 & 18,350 & 20.9\% \\
Politique & 7,406 & 3,140 & 10,546 & 12.0\% \\
Sport & 6,478 & 1,926 & 8,404 & 9.6\% \\
Économie & 5,955 & 2,219 & 8,174 & 9.3\% \\
Sciences \& Technologies & 1,982 & 2,215 & 4,197 & 4.8\% \\
\midrule
\textbf{Total} & \textbf{63,302} & \textbf{24,335} & \textbf{87,637} & \textbf{100\%} \\
\bottomrule
\addlinespace
\multicolumn{5}{@{}p{\columnwidth}@{}}{\footnotesize SHA-256 corpus fingerprint: \texttt{c14a8ae609d009c9289\allowbreak 393e6f0be761d\allowbreak da9f015aa4c17f515b\allowbreak 09388742a2c218}} \\
\end{tabular}
\caption{Dataset distribution by label source. Bucket A = deterministic slug rules; Bucket B = LLM-labeled ambiguous slugs.}
\label{tab:dataset_dist}
\end{table}

\subsection{Annotation Quality}
\label{sec:annotation_quality}

Quality assurance for Bucket A consists of verifying that the 74 slug rules map correctly to the intended harmonized categories, which we performed manually.

For Bucket B (27.8\%), we first establish that these labels are functionally inert for all reported results: training CamemBERT-base on the Bucket A training split alone (44,370 deterministic-slug articles) and evaluating on the identical held-out publisher pool yields a macro-F1 difference of $-0.004$ (95\% CI $[-0.024, +0.024]$; McNemar $p = 0.63$), with no per-category shift exceeding the noise band.

As independent quality evidence for these coverage labels, we conducted a four-rater agreement study on a 300-article stratified random sample drawn from ambiguous-slug sources (\texttt{/idees}, \texttt{/debats}, \texttt{/story}, \texttt{/flash-actu}, etc.), sampled proportionally to each slug type's share of Bucket B. The four raters are GPT-OSS-120B (original labels), Gemini-2.5-flash-lite (independent second-LLM pass, same seven-category prompt, temperature~$=0$), an adjudicator (who inspected inter-LLM disagreements and selected a label for each), and a blinded second human annotator (headline and body text only; no URLs, outlet names, or model labels).

The core quality signal is the blinded second human annotator (no knowledge of the URL-derived taxonomy, no model outputs, no role in adjudication), who independently recovers 84.0\% of the author's labels (95\% Wilson [79.4, 87.7]; Cohen's~$\kappa = 0.806$), bounding adjudication-specific bias to at most 16\% ($\pm 4.1$ points) of Bucket~B audit cases. Divergences cluster in the same boundary categories (\emph{Société}, \emph{Économie}) where the two LLMs also disagree. GPT-OSS agrees with the blinded human in 80.7\% of cases (95\% Wilson [75.8, 84.7]; $\kappa = 0.766$), a 3.3~pp gap below the human--human baseline, approaching inter-annotator agreement levels (Fleiss'~$\kappa = 0.835$ across all four raters).

Bucket B articles retain their LLM-assigned labels in the released dataset to ensure full corpus coverage for downstream users. Quality extrapolation beyond the 300-article audit (1.2\% of Bucket~B) assumes distributional representativeness of the stratified sample; the Bucket~A-only ablation above provides an independent bound on the impact of any residual labeling error.

The disagreement structure is informative. Across all rater pairs, disagreements concentrate in \emph{Société}, \emph{Culture \& Loisirs}, and \emph{Économie} (three-rater Krippendorff's~$\alpha = 0.789$), confirming that broad desk boundaries remain genuinely ambiguous even for humans, not solely an artifact of model labeling.

\subsection{Dataset Statistics}

The corpus is moderately imbalanced: \emph{Société} (22.1\%) and \emph{International} (21.3\%) together account for 43.4\% of articles, while \emph{Sciences \& Technologies} constitutes only 4.8\% (4,197 articles).

The dataset was split into training (70\%), validation (15\%), and test (15\%) sets using stratified random sampling with seed $= 42$, yielding 61,345 training, 13,146 validation, and 13,146 test articles. Text hashes were computed prior to splitting to prevent exact-duplicate leakage across splits. Because exact hashing can miss lightly edited syndicated copy, we additionally audited the 2,100-article publisher-held-out pool against the 61,345 training articles using rare 12-word lead shingles and full-text 5-gram containment. This audit found 0 exact normalized overlaps and 16/2,100 moderate-to-strong near-duplicate cases (containment $\geq 0.5$).

\section{Modeling and Experimental Setup}

\paragraph{Models.} We benchmark four trained classifiers and one prompting-based LLM comparator. The trained models are: (1) TF-IDF + Logistic Regression (character and word $n$-grams 1--3, L2 regularization) as a lexical baseline; (2) mBERT \citep{devlin2019}, a 12-layer multilingual BERT encoder; (3) CamemBERT-base \citep{martin2020}, a RoBERTa encoder pre-trained on \textasciitilde138GB of French text; (4) CamemBERTav2-base \citep{antoun2024}, a DeBERTaV3 encoder trained on 275B French tokens with a native 1,024-token context window. GPT-OSS-120B (\texttt{openai/gpt-oss-120b}; \citealp{openai2025gptoss}) is an open-weight, 117B-parameter Mixture-of-Experts model (5.1B active parameters per forward pass); it is evaluated under zero-shot and few-shot prompting on the held-out publisher pool in Section~\ref{sec:llm_comparison}. All transformer models use a linear classification head over [CLS] and are fine-tuned end-to-end (AdamW, lr $= 2\times10^{-5}$, batch 32, linear warmup, up to 5 epochs with early stopping patience $= 2$, checkpoint selected by best validation macro-F1, single NVIDIA L4 GPU). GPT-OSS-120B inference was run via AWS Bedrock (accessed May 2026); all other training used a single NVIDIA L4 GPU.

\paragraph{Input conditions.} Two conditions: headline + body (full text, 512-token truncation for 512-token models) and headline-only. Both conditions are evaluated for all four models.

\paragraph{Evaluation.} All models are trained on the 13-outlet FrenchNews-7 corpus (70/15/15 stratified split, seed $= 42$; $n_\text{test} = 13{,}146$). Cross-publisher evaluation uses a strictly disjoint held-out set of four outlets never seen during training. Macro-averaged F1 is the primary metric; per-class results are also reported. For the main test set and the pooled mixed-class held-out set, we report 95\% bootstrap confidence intervals over article-level predictions (1,000 replicates). For the held-out cross-publisher recall slices, we report 95\% Wilson intervals. To bound model instability from initialization, CamemBERT-base was fine-tuned on five random seeds (42, 123, 456, 789, 101112); per-seed cross-publisher macro-F1 ranges from 0.786 to 0.816 (CV $= 1.6\%$), and in-distribution variance is negligible (CV $= 0.36\%$).

Pairwise significance is assessed with paired bootstrap tests on macro-F1 and McNemar's test on per-article correctness, with Holm--Bonferroni correction within each evaluation family (Appendix~\ref{app:pairwise_significance}).

To ensure the LLM comparison is not artificially skewed by a zero-shot penalty, we also evaluate all three LLM comparators under few-shot prompting (5-shot and 7-shot balanced). Support examples are drawn strictly from the in-distribution validation split and restricted to deterministic Bucket A (URL-derived) labels only; no LLM-annotated examples serve as few-shot exemplars. This prevents both unseen-publisher leakage and circular reinforcement of the Bucket B label distribution.

\section{Results}

\subsection{Main Results}
\label{sec:main_results}

Table~\ref{tab:main_results} presents the full ablation results across all models and both input conditions (full text vs.\ headline-only) on the held-out test split ($n = 13{,}146$). Results reveal two findings: a clear performance hierarchy across model families, and a consistent advantage of full-text over headline-only input. For the recommended CamemBERT-base configuration, macro-F1 is 0.847 with a 95\% bootstrap CI of [0.841, 0.854], and accuracy is 0.860 with a 95\% bootstrap CI of [0.854, 0.866].

The $\star$ marks our recommended deployment configuration; $\dagger$ marks the best value per metric.

\begin{table*}[t]
\centering
\small
\setlength{\tabcolsep}{8pt}
\renewcommand{\arraystretch}{1.12}
\adjustbox{max width=\linewidth}{%
\begin{tabular}{@{}llccccc@{}}
\toprule
\textbf{Model} & \textbf{Input} & \textbf{Accuracy} & \textbf{Macro F1} & \textbf{Macro P} & \textbf{Macro R} & \textbf{$\Delta$F1 (H$-$F)} \\
\midrule
TF-IDF + LogReg & Headline only & 0.738 & 0.721 & 0.714 & 0.731 & N/A \\
TF-IDF + LogReg & Headline + body & 0.822 & 0.812 & 0.803 & 0.824 & $-$0.091 \\
mBERT & Headline only & 0.792 & 0.771 & 0.771 & 0.772 & N/A \\
mBERT & Headline + body & 0.845 & 0.833 & 0.835 & 0.831 & $-$0.062 \\
CamemBERTav2-base & Headline only & 0.813 & 0.797 & 0.796 & 0.797 & N/A \\
CamemBERTav2-base & Headline + body (512 tok) & 0.856 & 0.843 & 0.839 & 0.849 & $-$0.046 \\
CamemBERTav2-base & Headline + body$^{\ddagger}$ & 0.861 & 0.847 & 0.850 & 0.845 & $-$0.050 \\
CamemBERT-base & Headline only & 0.812 & 0.794 & 0.801 & 0.790 & N/A \\
CamemBERT-base & Headline + body $\star$ & 0.860$\dagger$ & 0.847$\dagger$ & 0.843$\dagger$ & 0.851$\dagger$ & $-$0.053 \\
\bottomrule
\end{tabular}}
\caption{Test set results ($n = 13{,}146$) across all models and input conditions. $\star$ = recommended deployment model; $\dagger$ = best value per metric (95\% bootstrap CIs for $\star$: Accuracy [0.854, 0.866], Macro F1 [0.841, 0.854]); $^{\ddagger}$evaluated at native 1,024-token context (all other rows use 512-token truncation). $\Delta$F1 (Headline only $-$ Full text): negative values indicate full text outperforms headline-only.}
\label{tab:main_results}
\end{table*}

Strong monolingual pretraining outperforms multilingual encoders on this task: the 1.4pp gap between mBERT and CamemBERT-base (95\% CI [$+0.009$, $+0.020$]) quantifies the value of French-specific pretraining. Under 512-token truncation, the 0.4pp gap between CamemBERT-base and CamemBERTav2 does not separate reliably (95\% CI [$-0.002$, $+0.009$]; McNemar $p = 0.147$). At its native 1,024-token context, CamemBERTav2 achieves macro-F1 = 0.847, matching CamemBERT-base identically and confirming that the truncation gap is an artifact of context-window limitation rather than architectural difference. We treat the two models as tied; CamemBERT-base remains the recommended deployment model.

\subsection{Per-Class Performance}

\begin{figure}[t]
\centering
\includegraphics[width=\columnwidth]{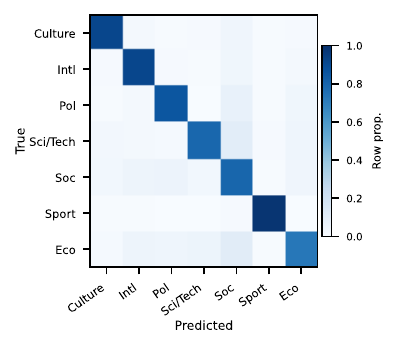}
\caption{Row-normalized confusion matrix for CamemBERT-base on the in-distribution test set ($n = 13{,}146$).}
\label{fig:confusion_in_dist}
\end{figure}

Figure~\ref{fig:confusion_in_dist} summarizes the in-distribution error profile of the recommended CamemBERT-base system. On the test set, \emph{Sport} achieves the highest F1 at 0.964, followed by \emph{Culture \& Loisirs} (0.924) and \emph{International} (0.902). The weakest class is \emph{Économie} (0.744, 95\% bootstrap CI [0.725, 0.763]), followed by \emph{Sciences \& Technologies} (0.753, 95\% bootstrap CI [0.726, 0.779]); \emph{Société} (0.796) also falls below the other four categories. The row-normalized confusion matrix shows \emph{Société} as the dominant error absorber, while \emph{Économie} leaks primarily into \emph{Société} and \emph{Politique}. This ordering broadly tracks category-boundary clarity and domain-specific vocabulary. Cross-publisher evaluation (\S\ref{sec:crosspub_gen}) confirms that \emph{Économie}'s difficulty reflects editorial boundary ambiguity rather than recoverable headroom: blinded human agreement on held-out \emph{Économie} articles reaches only 55\% (22/40; 95\% Wilson [39.8, 69.3]) --- an interval wide enough to contain the classifier's 0.517 cross-publisher recall, so on this sample the two are statistically indistinguishable rather than the model demonstrably reaching a human ceiling.

\subsection{Full-Text vs.\ Headline-Only Input}

Full-text input consistently outperforms headline-only across all four model families (Table~\ref{tab:main_results}, $\Delta$F1 column). The gap is largest for TF-IDF and smallest for CamemBERTav2, indicating that contextual representations are less dependent on full text than bag-of-words features. Headline-only results are provided as a deployment-constrained fallback (e.g., RSS feeds, paywalled archives).

\subsection{Out-of-Distribution (OOD) Generalization Across Publishers}
\label{sec:crosspub_gen}

\begin{figure}[t]
\centering
\includegraphics[width=0.66\columnwidth]{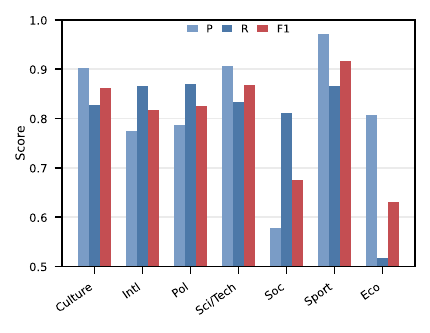}
\caption{Per-class precision, recall, and F1 on the pooled mixed-class cross-publisher test set ($n = 2{,}100$, 300 per category).}
\label{fig:crosspub_mixed_perclass}
\end{figure}

Four outlets never seen during training (\emph{Libération}, \emph{BFMTV}, \emph{Le Nouvel Obs}, and \emph{Franceinfo}) were designated as held-out evaluation targets prior to model training, mirroring the format diversity of the training corpus (a national daily, a private broadcaster, a weekly magazine, and a public broadcaster); they were collected separately and are not part of the 87,637-article corpus. These four outlets provide a balanced mixed-class test set of 2,100 articles (300 per category, 7-way classification), with publisher URL slugs mapped into the FrenchNews-7 taxonomy by the same 74-rule lookup table used for Bucket~A, which resolves all 2,100 held-out URLs with no outlet-specific additions. A blind annotation of 270 held-out articles (10 per outlet--category cell; \emph{BFMTV} $\times$ \emph{Sport} was skipped as no Sport articles were available, giving $4 \times 7 \times 10 - 10 = 270$; same protocol as \S\ref{sec:annotation_quality}, no URL or outlet shown) recovered publisher-assigned labels in 82.6\% of cases ($\kappa = 0.80$); the sole exception was \emph{Économie} ($55\%$), illustrating the gap between external semantic intuition and editorial institutional discretion for political-economy content. This is consistent with the benchmark's design: editorial boundary ambiguity appears to be a task property rather than a labeling artifact, particularly for \emph{Économie}. On this pooled unseen-outlet test, CamemBERT-base reaches accuracy $= 0.799$ (95\% bootstrap CI [0.782, 0.816]) and macro-F1 $= 0.799$ (95\% bootstrap CI [0.783, 0.815]) (macro-precision $= 0.817$). CamemBERTav2 at its native 1,024-token context achieves identical cross-publisher performance (macro-F1 $= 0.798$, $\Delta < 0.001$; see Appendix~\ref{app:full_context}). Excluding 16 flagged near-duplicate cases leaves performance essentially unchanged ($\Delta$ macro-F1 $< 0.001$). Figure~\ref{fig:crosspub_mixed_perclass} shows the per-class transfer profile.

\emph{Économie} remains the weakest class under shift (precision $= 0.807$, recall $= 0.517$, F1 $= 0.630$), while \emph{Sport} remains strongest (F1 $= 0.915$). \emph{Société} acts as an error absorber under distribution shift: its recall is 0.810, but precision falls to 0.577 because the model defaults to this broad sink category for boundary-ambiguous texts. Binary per-category recall slices confirm the pattern: six of seven categories achieve recall $\geq 0.810$ (e.g., \emph{Politique} 0.870 [0.827, 0.903]; \emph{International} 0.867 [0.824, 0.901]), while \emph{Économie} remains the weak point at 0.517 [0.460, 0.573]. Blinded human agreement with \emph{Économie} slugs is 55\% on the same outlets (22/40; 95\% Wilson [39.8, 69.3]), and the classifier's 0.517 recall falls inside that interval. At $n = 40$ the comparison cannot separate the two; we therefore read this as consistent with editorial routing conventions driving the gap rather than as evidence that classifier headroom is exhausted. The most frequent held-out errors are \emph{Économie} $\rightarrow$ \emph{Société} (64 cases), \emph{Économie} $\rightarrow$ \emph{Politique} (37), and \emph{Sciences \& Technologies} $\rightarrow$ \emph{Société} (29). The held-out pool is balanced at 300 articles per category, which favours classes with structurally clean slugs. The temporal check in Appendix~\ref{app:case_study} is less favourable on the same classes: accuracy 82.1\%, per-class recall 57.2\% (\emph{Économie}) to 96.9\% (\emph{Sport}), with \emph{Sciences \& Technologies} at 59.8\% against 0.833 here (Table~\ref{tab:longitudinal_slug_check}).

\paragraph{Impact of publisher imbalance.}
A potential confound in the cross-publisher evaluation is the heavy representation of \emph{Le Monde} in the training data (37.8\%), which risks the model learning a publisher's house style rather than publisher-agnostic editorial boundaries. To test this, we trained an ablation model in which \emph{Le Monde} was aggressively downsampled to match the next largest publisher (\emph{L'Express}, $n = 6{,}377$). Evaluating this balanced model on the same unseen-outlet pool yields performance that is statistically indistinguishable from the full baseline (macro-F1 $0.802$ vs.\ $0.799$; $\Delta = {+}0.003$, 95\% CI $[-0.019, {+}0.027]$, McNemar $p = 0.659$). This provides evidence against the hypothesis that the classifier relies on memorising \emph{Le Monde}'s house style; the model successfully learns publisher-agnostic boundaries even when the training distribution is artificially balanced. Feature attribution via Integrated Gradients corroborates this: top-scoring tokens are topically grounded across all seven classes, with no publisher-identifying surface marker appearing in any top-10 attribution list (Appendix~\ref{app:publisher_analysis}).

\subsection{Comparison with Zero-Shot LLM Performance}
\label{sec:llm_comparison}

We compare fine-tuned CamemBERT-base against GPT-OSS-120B on the OOD test set, scored as seven one-vs-rest recall slices. The prompting baseline uses the same seven category definitions in three settings: zero-shot, 5-shot, and 7-shot balanced.

Fine-tuned CamemBERT-base exceeds GPT-OSS-120B across all prompt settings. GPT-OSS-120B zero-shot achieves mean accuracy $= 0.753$ and macro-F1 $= 0.758$ (three seeds) on the 2,100-article unseen-outlet pool. Mistral Small~3.2 \citep{mistral2025small32} reaches $0.772/0.775$ and Llama-3.3-70B-Instruct \citep{meta2024llama33} reaches $0.778/0.771$ (accuracy/macro-F1). All three LLMs cluster below fine-tuned CamemBERT-base ($0.799$), though Llama-3.3-70B leads on \emph{Politique} (0.907) and \emph{International} (0.930). Balanced few-shot prompting improves none of the three comparators and changes no ranking (Table~\ref{tab:fewshot}); the strongest single configuration (GPT-OSS-120B, 5-shot, seed~11: $0.758/0.761$) remains below CamemBERT-base. On a seven-way desk-routing task the label is a publisher routing convention rather than a pattern a handful of demonstrations conveys, so exemplars add noise rather than signal. Table~\ref{tab:llm_comparison} shows per-category zero-shot means. CamemBERT's margin comes from \emph{Société} and \emph{Économie}, where editorial boundary conventions are least recoverable from short prompts.

\begin{table}[t]
\centering
\small
\setlength{\tabcolsep}{4pt}
\renewcommand{\arraystretch}{1.08}
\begin{adjustbox}{max width=\columnwidth}
\begin{tabular}{@{}lrrrrr@{}}
\toprule
Category & CamemBERT & Bucket-A & GPT-OSS & Mistral & Llama-3.3 \\
\midrule
Politique & 0.870 & 0.827 & 0.848 & 0.843 & \textbf{0.907} \\
International & 0.867 & 0.873 & 0.777 & 0.783 & \textbf{0.930} \\
Culture \& Loisirs & 0.827 & 0.827 & 0.821 & 0.830 & \textbf{0.847} \\
Sport & 0.867 & 0.860 & 0.756 & 0.870 & \textbf{0.883} \\
Sciences \& Tech & 0.833 & 0.783 & 0.832 & \textbf{0.847} & 0.820 \\
Société & \textbf{0.810} & 0.807 & 0.776 & 0.780 & 0.673 \\
Économie & \textbf{0.517} & 0.583 & 0.462 & 0.453 & 0.383 \\
\midrule
Avg recall & \textbf{0.799} & 0.794 & 0.753 & 0.772 & 0.778 \\
\bottomrule
\end{tabular}
\end{adjustbox}
\caption{Per-category recall on the 2,100-article unseen-outlet pool. GPT-OSS = three-seed mean; Mistral = Mistral Small~3.2; Llama-3.3 = Llama-3.3-70B-Instruct; Bucket-A = URL-derived labels only. Few-shot results for all three LLMs are in Table~\ref{tab:fewshot}.}
\label{tab:llm_comparison}
\end{table}

\begin{table}[t]
\centering
\small
\setlength{\tabcolsep}{5pt}
\renewcommand{\arraystretch}{1.08}
\begin{tabular}{@{}lccc@{}}
\toprule
Model & 0-shot & 5-shot & 7-shot \\
\midrule
GPT-OSS-120B & 0.758 & 0.755 & 0.756 \\
Mistral$^{\dagger}$ & 0.774 & 0.767 & 0.764 \\
Llama-3.3-70B & 0.771 & 0.770 & 0.761 \\
\midrule
CamemBERT-base (fine-tuned) & \textbf{0.799} & --- & --- \\
\bottomrule
\end{tabular}
\caption{Macro-F1 on the 2,100-article unseen-outlet pool under balanced few-shot prompting (three seeds per setting; support examples drawn only from deterministic Bucket~A validation items, so no LLM-annotated or unseen-publisher item is used as an exemplar). Few-shot improves no comparator and changes no ranking. $^{\dagger}$Mistral Small~3.2 was no longer served by the provider when the few-shot runs were carried out; this row uses Mistral-Small-4-119B, whose re-run 0-shot macro-F1 on the identical harness is 0.774, within 0.001 of Mistral Small~3.2's 0.775 (the value in Table~\ref{tab:llm_comparison}).}
\label{tab:fewshot}
\end{table}

\section{Discussion}

To our knowledge, FrenchNews-7 is the first French benchmark combining an empirically derived multi-publisher taxonomy, publisher-native URL labels, LLM annotation for ambiguous cases, and held-out cross-publisher evaluation across all seven categories. The taxonomy is reproducible for other media ecosystems via the same URL-slug derivation (IPTC convergence: Appendix~\ref{app:iptc_crosswalk}). Under pooled mixed-class evaluation on unseen outlets, CamemBERT-base sustains macro-F1 $\approx 0.80$ when discriminating among all seven classes, indicating that URL-derived labels yield a reusable classifier rather than a narrow per-category recognizer. Appendix~\ref{app:case_study} adds a complementary temporal-generalization check on later articles from the same 13 publishers (\S\ref{sec:crosspub_gen}) and a scale demonstration on a 2-million-article longitudinal corpus, revealing distinct editorial fingerprints and growing topical differentiation across French outlets (2022--2025).

Under distribution shift, \emph{Société} absorbs boundary-ambiguous text while \emph{Économie}'s difficulty is rooted in cross-publisher editorial discretion rather than recoverable headroom (\S\ref{sec:crosspub_gen}, \S\ref{sec:annotation_quality}); the release prioritises robustness and reproducibility over boundary precision.

\section*{Limitations}

The seven-class taxonomy is deliberately coarse; finer distinctions such as \emph{Environnement} and \emph{Santé} are collapsed into broader classes. Three convergent lines of evidence in Appendix~\ref{app:publisher_analysis} argue against publisher memorisation (per-publisher F1 breakdown, Integrated Gradients attribution, and publisher-token masking). Cross-publisher evaluation includes both a balanced mixed-class 7-way test and per-category recall slices on four news-format held-out outlets, while Appendix~\ref{app:case_study} adds a large-n temporal check on later articles from the same 13 publishers.

The 2,100-article mixed-class held-out test uses publisher URL slugs as reference labels, consistent with the operational definition of the task established in \S\ref{sec:crosspub_gen}: the task is to predict editorial desk labels, and publisher-assigned URL slugs are the reference labels by construction. The blinded human audit (\S\ref{sec:annotation_quality}) does not correct these reference labels but demonstrates that they carry recoverable semantic signal. Bucket B is spot-checked on 300 ambiguous-URL articles; the remaining 24,035 are unaudited. The four-rater agreement study (\S\ref{sec:annotation_quality}) provides quality evidence through symmetric multi-rater analysis, but the 24,335 Bucket~B training labels are single-LLM outputs with no per-article human verification; scaling multi-annotator consensus across this subset remains future work.

Temporal generalization beyond the 2022--2025 window and outside the 13 known publishers remains unverified. All 13 outlets are France-based; the taxonomy has not been validated on francophone media from Belgium, Switzerland, Quebec, or Africa, where distinct editorial conventions may yield different category boundaries. We report bootstrap confidence intervals for the main test-set metrics and Wilson intervals for the held-out recall slices; these quantify test-sample uncertainty only.

The classifier's utility varies substantially by downstream research task. Appendix~\ref{app:reliability_tiers} maps three reliability tiers to concrete task types based on the cross-publisher evaluation results.

\section*{Ethics Statement}

\ifarxiv
The fine-tuned CamemBERT-base model and the labeled manifest dataset are publicly released on Hugging Face (\url{https://huggingface.co/LeFrenchNewsLab/camembert-base-frenchnews7} and \url{https://huggingface.co/datasets/LeFrenchNewsLab/frenchnews-7}).
\else
The fine-tuned CamemBERT-base model and the labeled manifest dataset are available at [anonymous link].
\fi
The manifest contains article-level metadata (URL, publication date, publisher, assigned category, bucket provenance, and SHA-256 text hash) but does not redistribute verbatim article text or full headlines due to copyright restrictions. Researchers can inspect labels, metadata, and evaluation artifacts directly; authorized users may re-fetch source pages under their own legal and institutional frameworks using the provided reference collection scripts. The 74-rule slug lookup table and seven-category prompt template are included in the dataset repository. The model is released as open weights with a model card documenting training configuration, intended use, and per-class reliability tiers (Appendix~\ref{app:reliability_tiers}).

The benchmark is designed for cross-publisher editorial desk classification, not for person-level inference or fully general claims about French media structure. Annotation procedures are described in full in \S\ref{sec:annotation_pipeline}--\ref{sec:annotation_quality}. Remaining risks include correlated LLM labeling errors and residual outlet-style bias. AI assisted prose editing, analysis-code debugging, and manuscript review.

\ifnldp
  \bibliographystyle{unsrtnat}
\fi
\bibliography{paper_refs}

\begin{thebibliography}{27}
\providecommand{\natexlab}[1]{#1}

\bibitem[{Antoun et~al.(2024)Antoun, Kulumba, Touchent, de~la Clergerie, Sagot, and Seddah}]{antoun2024}
Wissam Antoun, Francis Kulumba, Rian Touchent, {\'E}ric de~la Clergerie, Beno{\^i}t Sagot, and Djam{\'e} Seddah. 2024.
\newblock \href {https://doi.org/10.48550/arXiv.2411.08868} {Camembert 2.0: A smarter french language model aged to perfection}.
\newblock \emph{arXiv preprint arXiv:2411.08868}.

\bibitem[{Devlin et~al.(2019)Devlin, Chang, Lee, and Toutanova}]{devlin2019}
Jacob Devlin, Ming-Wei Chang, Kenton Lee, and Kristina Toutanova. 2019.
\newblock \href {https://doi.org/10.18653/v1/N19-1423} {Bert: Pre-training of deep bidirectional transformers for language understanding}.
\newblock In \emph{Proceedings of NAACL-HLT 2019}, pages 4171--4186.

\bibitem[{Ding et~al.(2023)Ding, Qin, Liu, Chia, Li, Joty, and Bing}]{ding2023}
Bosheng Ding, Chengwei Qin, Linlin Liu, Yew~Ken Chia, Boyang Li, Shafiq Joty, and Lidong Bing. 2023.
\newblock \href {https://doi.org/10.18653/v1/2023.acl-long.626} {Is gpt-3 a good data annotator?}
\newblock In \emph{Proceedings of the 61st Annual Meeting of the Association for Computational Linguistics (Volume 1: Long Papers)}, pages 11173--11195.

\bibitem[{Escouflaire et~al.(2024)Escouflaire, Descampe, and Fairon}]{escouflaire2024}
Louis Escouflaire, Antonin Descampe, and C{\'e}drick Fairon. 2024.
\newblock \href {https://doi.org/10.1016/j.langcom.2024.09.004} {Automated text classification of opinion vs.\ news {French} press articles: {A} comparison of transformer and feature-based approaches}.
\newblock \emph{Language \& Communication}, 99:129--140.

\bibitem[{Field et~al.(2018)Field, Kliger, Wintner, Pan, Jurafsky, and Tsvetkov}]{field2018}
Anjalie Field, Doron Kliger, Shuly Wintner, Jennifer Pan, Dan Jurafsky, and Yulia Tsvetkov. 2018.
\newblock \href {https://doi.org/10.18653/v1/D18-1393} {Framing and agenda-setting in {R}ussian news: a computational analysis of intricate political strategies}.
\newblock In \emph{Proceedings of the 2018 Conference on Empirical Methods in Natural Language Processing}, pages 3570--3580. Association for Computational Linguistics.

\bibitem[{Gilardi et~al.(2023)Gilardi, Alizadeh, and Kubli}]{gilardi2023}
Fabrizio Gilardi, Meysam Alizadeh, and Ma{\"e}l Kubli. 2023.
\newblock \href {https://doi.org/10.1073/pnas.2305016120} {{ChatGPT} outperforms crowd workers for text-annotation tasks}.
\newblock \emph{Proceedings of the National Academy of Sciences}, 120(30):e2305016120.

\bibitem[{He et~al.(2021)He, Liu, Gao, and Chen}]{he2021}
Pengcheng He, Xiaodong Liu, Jianfeng Gao, and Weizhu Chen. 2021.
\newblock \href {https://doi.org/10.48550/arXiv.2006.03654} {Deberta: Decoding-enhanced bert with disentangled attention}.
\newblock In \emph{Proceedings of ICLR 2021}.

\bibitem[{Icard et~al.(2024)Icard, Maine, Casanova, Faye, Chanson, Gadek, Atemezing, Bancilhon, and \'Egr\'e}]{icard2024}
Benjamin Icard, Fran\c{c}ois Maine, Morgane Casanova, G\'eraud Faye, Julien Chanson, Guillaume Gadek, Ghislain Atemezing, Fran\c{c}ois Bancilhon, and Paul \'Egr\'e. 2024.
\newblock A multi-label dataset of {F}rench fake news: Human and machine insights.
\newblock In \emph{Proceedings of LREC-COLING 2024}, pages 812--818, Torino, Italia. ELRA and ICCL.

\bibitem[{{International Press Telecommunications Council}(2025)}]{iptc_mediatopics_2025}
{International Press Telecommunications Council}. 2025.
\newblock {IPTC Media Topics}.
\newblock \url{https://iptc.org/standards/media-topics/}.
\newblock Accessed: 2026-05-09.

\bibitem[{Joachims(1998)}]{joachims1998}
Thorsten Joachims. 1998.
\newblock \href {https://doi.org/10.1007/BFb0026683} {Text categorization with support vector machines: Learning with many relevant features}.
\newblock In \emph{Proceedings of ECML 1998}, pages 137--142.

\bibitem[{Kuzman and Ljube{\v{s}}i{\'c}(2025)}]{kuzman2025}
Taja Kuzman and Nikola Ljube{\v{s}}i{\'c}. 2025.
\newblock \href {https://doi.org/10.1109/ACCESS.2025.3544814} {Llm teacher-student framework for text classification with no manually annotated data: A case study in iptc news topic classification}.
\newblock \emph{IEEE Access}, 13:35621--35633.
\newblock ArXiv:2411.19638.

\bibitem[{Le et~al.(2020)Le, Vial, Frej, Segonne, Coavoux, Lecouteux, Allauzen, Crabb{\'e}, Besacier, and Schwab}]{le2020}
Hang Le, Lo{\"i}c Vial, Jibril Frej, Vincent Segonne, Maximin Coavoux, Benjamin Lecouteux, Alexandre Allauzen, Beno{\^i}t Crabb{\'e}, Laurent Besacier, and Didier Schwab. 2020.
\newblock \href {https://doi.org/10.48550/arXiv.1912.05372} {Flaubert: Unsupervised language model pre-training for french}.
\newblock In \emph{Proceedings of LREC 2020}, pages 2479--2490.

\bibitem[{Lewis et~al.(2004)Lewis, Yang, Rose, and Li}]{lewis2004}
David~D. Lewis, Yiming Yang, Tony~G. Rose, and Fan Li. 2004.
\newblock Rcv1: A new benchmark collection for text categorization research.
\newblock \emph{Journal of Machine Learning Research}, 5:361--397.

\bibitem[{Liu et~al.(2019)Liu, Ott, Goyal, Du, Joshi, Chen, Levy, Lewis, Zettlemoyer, and Stoyanov}]{liu2019}
Yinhan Liu, Myle Ott, Naman Goyal, Jingfei Du, Mandar Joshi, Danqi Chen, Omer Levy, Mike Lewis, Luke Zettlemoyer, and Veselin Stoyanov. 2019.
\newblock \href {https://doi.org/10.48550/arXiv.1907.11692} {Roberta: A robustly optimized bert pretraining approach}.
\newblock \emph{arXiv preprint arXiv:1907.11692}.

\bibitem[{Lu and Smith(2025)}]{lu2025}
Yucheng Lu and Kazimier Smith. 2025.
\newblock \href {https://doi.org/10.48550/arXiv.2504.15432} {Feeding llm annotations to bert classifiers at your own risk}.
\newblock \emph{arXiv preprint arXiv:2504.15432}.
\newblock Preprint under review.

\bibitem[{Martin et~al.(2020)Martin, Muller, Su{\'a}rez, Dupont, Romary, de~la Clergerie, Seddah, and Sagot}]{martin2020}
Louis Martin, Benjamin Muller, Pedro Javier~Ortiz Su{\'a}rez, Yoann Dupont, Laurent Romary, {\'E}ric~Villemonte de~la Clergerie, Djam{\'e} Seddah, and Beno{\^i}t Sagot. 2020.
\newblock \href {https://doi.org/10.18653/v1/2020.acl-main.645} {Camembert: A tasty french language model}.
\newblock In \emph{Proceedings of ACL 2020}, pages 7203--7219.

\bibitem[{McCombs and Shaw(1972)}]{mccombs1972}
Maxwell~E. McCombs and Donald~L. Shaw. 1972.
\newblock \href {https://doi.org/10.1086/267990} {The agenda-setting function of mass media}.
\newblock \emph{Public Opinion Quarterly}, 36(2):176--187.

\bibitem[{{Meta AI}(2024)}]{meta2024llama33}
{Meta AI}. 2024.
\newblock Llama 3.3 70{B} instruct.
\newblock \url{https://huggingface.co/meta-llama/Llama-3.3-70B-Instruct}.
\newblock Accessed: 2026-05-26.

\bibitem[{{Mistral AI}(2025)}]{mistral2025small32}
{Mistral AI}. 2025.
\newblock Mistral small 3.2.
\newblock \url{https://huggingface.co/mistralai/Mistral-Small-3.2-24B-Instruct-2506}.
\newblock Accessed: 2026-05-26.

\bibitem[{{OpenAI}(2025)}]{openai2025gptoss}
{OpenAI}. 2025.
\newblock \href {https://doi.org/10.48550/arXiv.2508.10925} {{gpt-oss-120b \& gpt-oss-20b Model Card}}.
\newblock Technical report, OpenAI.
\newblock ArXiv:2508.10925.

\bibitem[{Pangakis and Wolken(2024)}]{pangakisw2024acl}
Nicholas Pangakis and Sam Wolken. 2024.
\newblock \href {https://doi.org/10.18653/v1/2024.nlpcss-1.9} {Knowledge distillation in automated annotation: Supervised text classification with llm-generated training labels}.
\newblock In \emph{Proceedings of the Sixth Workshop on Natural Language Processing and Computational Social Science (NLP+CSS 2024)}, pages 113--131.

\bibitem[{Pelloin et~al.(2024)Pelloin, Dodson, Chapuis, Herv{\'{e}}, and Doukhan}]{pelloin2024}
Valentin Pelloin, Lena Dodson, Emile Chapuis, Nicolas Herv{\'{e}}, and David Doukhan. 2024.
\newblock \href {https://doi.org/10.21437/Interspeech.2024-1854} {Automatic classification of news subjects in broadcast news: Application to a gender bias representation analysis}.
\newblock In \emph{Proceedings of Interspeech 2024}, pages 3055--3059.
\newblock ArXiv:2407.14180.

\bibitem[{Scialom et~al.(2020)Scialom, Dray, Lamprier, Piwowarski, and Staiano}]{scialom2020}
Thomas Scialom, Paul-Alexis Dray, Sylvain Lamprier, Benjamin Piwowarski, and Jacopo Staiano. 2020.
\newblock \href {https://doi.org/10.18653/v1/2020.emnlp-main.647} {Mlsum: The multilingual summarization corpus}.
\newblock In \emph{Proceedings of EMNLP 2020}, pages 8051--8067.

\bibitem[{Sundararajan et~al.(2017)Sundararajan, Taly, and Yan}]{sundararajan2017}
Mukund Sundararajan, Ankur Taly, and Qiqi Yan. 2017.
\newblock Axiomatic attribution for deep networks.
\newblock In \emph{Proceedings of the 34th International Conference on Machine Learning}, ICML 2017, pages 3319--3328.

\bibitem[{T{\"o}rnberg(2023)}]{tornberg2023}
Petter T{\"o}rnberg. 2023.
\newblock \href {https://doi.org/10.48550/arXiv.2304.06588} {{ChatGPT-4} outperforms experts and crowd workers in annotating political {Twitter} messages with zero-shot learning}.
\newblock \emph{arXiv preprint arXiv:2304.06588}.

\bibitem[{Wastl et~al.(2025)Wastl, Vamvas, Calleri, and Sennrich}]{wastl2025}
Michelle Wastl, Jannis Vamvas, Selena Calleri, and Rico Sennrich. 2025.
\newblock 20min-{XD}: A comparable corpus of {S}wiss news articles.
\newblock In \emph{Proceedings of SwissText 2025}.
\newblock ArXiv:2504.21677.

\bibitem[{Zhang et~al.(2015)Zhang, Zhao, and LeCun}]{zhang2015}
Xiang Zhang, Junbo Zhao, and Yann LeCun. 2015.
\newblock Character-level convolutional networks for text classification.
\newblock In \emph{Advances in Neural Information Processing Systems 28}, pages 649--657.

\end{thebibliography}

\appendix

\section{Publisher-Level Analysis and Feature Attribution}
\label{app:publisher_analysis}

\paragraph{Per-publisher test performance.}
Per-publisher macro-F1 on the in-distribution test set spans from 0.626 (\emph{TF1 INFO}) to 0.906 (\emph{Ouest-France}), a range of 0.280 points across 13 publishers (Table~\ref{tab:publisher_breakdown}). Spearman~$\rho = 0.088$ ($p = 0.775$, $N = 13$) between each publisher's training-set size and their test macro-F1 indicates no statistically significant size--performance correlation; this result should be treated as indicative rather than inferential given the small $N$. \emph{Le Monde}, the largest training contributor (37.8\%, $N_\text{test} = 5{,}010$), achieves macro-F1 $= 0.849$, within one standard deviation of the cross-publisher mean ($\bar{x} = 0.795$) and therefore not an outlier. Together with the Le Monde balanced ablation (Section~\ref{sec:crosspub_gen}), these results argue against the publisher-memorisation hypothesis. \emph{Sciences \& Technologies} shows the highest per-publisher variance, consistent with its low overall support ($N = 667$), and should be interpreted cautiously for small-N outlets (La Croix $N = 272$; TF1~INFO $N = 291$). Zero-F1 cells reflect categories with no test articles for that publisher (JDD Sciences~\& Technologies: $N=0$; TF1~INFO Sport: $N=0$).

\begin{table*}[h]
\centering
\small
\setlength{\tabcolsep}{4pt}
\renewcommand{\arraystretch}{1.05}
\begin{adjustbox}{max width=\textwidth}
\begin{tabular}{lrrrrrrrrrr}
\toprule
Publisher & N & Acc. & Macro-F1 & Cult. & Éco. & Intl. & Pol. & Sci\&T. & Soc. & Sport \\
\midrule
20 Minutes & 329 & 0.851 & 0.752 & 0.870 & 0.615 & 0.899 & 0.835 & 0.250 & 0.838 & 0.958 \\
JDD & 1055 & 0.939 & 0.691 & 0.977 & 0.316 & 0.899 & 0.859 & 0.000 & 0.809 & 0.977 \\
L'Express & 1368 & 0.816 & 0.797 & 0.817 & 0.737 & 0.905 & 0.780 & 0.635 & 0.799 & 0.909 \\
L'Humanité & 1209 & 0.843 & 0.831 & 0.911 & 0.717 & 0.899 & 0.835 & 0.696 & 0.782 & 0.979 \\
La Croix & 272 & 0.871 & 0.806 & 0.923 & 0.444 & 0.892 & 0.793 & 0.750 & 0.850 & 0.988 \\
Le Figaro & 1122 & 0.877 & 0.832 & 0.883 & 0.757 & 0.894 & 0.815 & 0.726 & 0.749 & 0.999 \\
Le HuffPost & 382 & 0.935 & 0.898 & 0.928 & 0.800 & 0.807 & 0.975 & 0.889 & 0.919 & 0.966 \\
\textbf{Le Monde} & \textbf{5010} & \textbf{0.858} & \textbf{0.849} & \textbf{0.933} & \textbf{0.779} & \textbf{0.908} & \textbf{0.821} & \textbf{0.768} & \textbf{0.788} & \textbf{0.943} \\
Le Parisien & 345 & 0.884 & 0.763 & 0.883 & 0.444 & 0.896 & 0.880 & 0.400 & 0.857 & 0.980 \\
Le Point & 1174 & 0.819 & 0.791 & 0.866 & 0.657 & 0.897 & 0.838 & 0.610 & 0.773 & 0.898 \\
Ouest-France & 277 & 0.931 & 0.906 & 0.951 & 0.821 & 0.968 & 0.935 & 0.769 & 0.899 & 1.000 \\
Slate.fr & 312 & 0.804 & 0.795 & 0.791 & 0.625 & 0.867 & 0.989 & 0.752 & 0.741 & 0.800 \\
TF1 INFO & 291 & 0.842 & 0.626 & 0.826 & 0.167 & 0.863 & 0.861 & 0.910 & 0.757 & 0.000 \\
\midrule
\textit{All (pooled)} & 13146 & 0.860 & 0.847 & 0.924 & 0.744 & 0.902 & 0.843 & 0.753 & 0.796 & 0.964 \\
\bottomrule
\end{tabular}
\end{adjustbox}
\caption{Per-publisher performance on the in-distribution 13-publisher test set ($N = 13{,}146$). \textbf{Le Monde} bolded as the dominant training contributor (37.8\%). Cult.\ = Culture \& Loisirs; Éco.\ = Économie; Intl.\ = International; Pol.\ = Politique; Sci\&T.\ = Sciences \& Technologies; Soc.\ = Société.}
\label{tab:publisher_breakdown}
\end{table*}

\begin{figure*}[t]
\centering
\includegraphics[width=\textwidth]{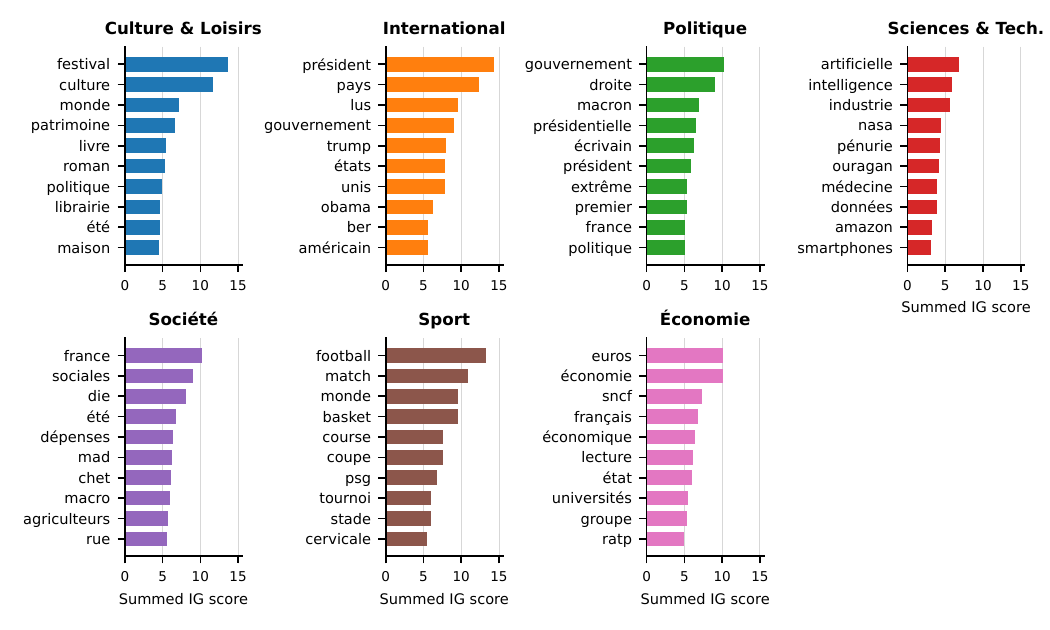}
\caption{Top-10 tokens per class by aggregated Integrated Gradients attribution score (199-article stratified sample, CamemBERT-base full-text model). All leading tokens are topically interpretable; no publisher-identifying surface marker appears in any top-10 list.}
\label{fig:top_tokens}
\end{figure*}

\paragraph{Feature attribution via Integrated Gradients.}
To verify that the model attends to topical vocabulary rather than publisher-identifying surface cues, we ran Integrated Gradients \citep[IG;][]{sundararajan2017} on a stratified 199-article sample from the in-distribution test split (minimum 2 per publisher; proportional stratification within publisher and class; seed 42). We used the \texttt{captum} library with a zero-embedding baseline and 50 integration steps; attribution scores were aggregated at the subword-token level (L1 norm over the embedding dimension) and pooled per class by summing across all gold-matching articles. Figure~\ref{fig:top_tokens} shows the top-10 tokens per class. Across all seven classes, leading tokens are unambiguously topical: \emph{festival}, \emph{culture}, \emph{patrimoine} for Culture~\&~Loisirs; \emph{président}, \emph{gouvernement}, \emph{macron} for Politique; \emph{football}, \emph{match}, \emph{psg} for Sport; \emph{artificielle}, \emph{intelligence}, \emph{nasa} for Sciences~\&~Technologies; \emph{euros}, \emph{sncf}, \emph{économie} for Économie. No publisher domain stem, section header, or byline fragment appears in any top-10 list.

\paragraph{Publisher-token masking experiment.}
As a complementary check, we applied a conservative masking set to the same 199 articles before re-running inference. Three token categories were masked (replaced with a blank token): (1) bylines (\textit{Par Prénom Nom} following a newline, regex-detected), (2) publisher section headers (fixed string list, case-insensitive), and (3) publisher domain stems (\texttt{lemonde.fr}, \texttt{lefigaro.fr}, etc., regex). Dateline cities were excluded to avoid inadvertently masking topical signal (e.g., \textit{Washington}~$\to$ International). Across 199~articles, only 4~tokens matched the masking criteria (1~byline, 1~section header, 2~domain substrings), representing 0.004\% of the body-token budget. Table~\ref{tab:masking_results} shows that aggregate $\Delta$\,F1 is 0.000 with no per-class change. Publisher-identifying surface tokens are effectively absent from article bodies in this corpus; removing them has no effect on predictions, a directional result consistent with the per-publisher breakdown and the Le Monde balanced ablation (Section~\ref{sec:crosspub_gen}), though the small number of masked tokens limits its evidentiary weight.

\begin{table}[h]
\centering
\small
\setlength{\tabcolsep}{5pt}
\renewcommand{\arraystretch}{1.05}
\begin{adjustbox}{max width=\columnwidth}
\begin{tabular}{lrrr}
\toprule
Category & Baseline F1 & Masked F1 & $\Delta$ \\
\midrule
Culture \& Loisirs & 0.930 & 0.930 & 0.000 \\
Économie & 0.723 & 0.723 & 0.000 \\
International & 0.909 & 0.909 & 0.000 \\
Politique & 0.760 & 0.760 & 0.000 \\
Sciences \& Technologies & 0.710 & 0.710 & 0.000 \\
Société & 0.744 & 0.744 & 0.000 \\
Sport & 0.955 & 0.955 & 0.000 \\
\midrule
Macro (all) & 0.819 & 0.819 & 0.000 \\
\bottomrule
\end{tabular}
\end{adjustbox}
\caption{Per-class F1 before and after publisher-token masking (199-article sample, CamemBERT-base). Only 4 tokens were masked across all articles (0.004\% of body-token budget); no class changes by any amount.}
\label{tab:masking_results}
\end{table}

\section{IPTC Media Topics Crosswalk}
\label{app:iptc_crosswalk}

FrenchNews-7 is derived observationally from French newsroom routing URLs rather than a theoretical ontology; it deliberately collapses several fine-grained distinctions made by IPTC. For example, environmental coverage (\textit{Environnement}) is distributed across \textit{Société} and \textit{Sciences \& Technologies}, while \textit{Santé} (Health) is folded into \textit{Société}.

Of the 17 top-level IPTC Media Topics categories, 10 map to at least one FrenchNews-7 class (58.8\% structural overlap). The remaining seven have no straightforward FrenchNews-7 equivalent: Disaster/Accident (03000000), Environment (06000000), Human Interest (08000000), Labour (09000000), Religion (12000000), Conflict/War/Peace (16000000), and Weather (17000000). \textit{Société} is the largest collision point, absorbing four IPTC categories (Society, Crime/Law/Justice, Education, Health). Table~\ref{tab:iptc_mapping} provides the full crosswalk with IPTC numeric codes and per-class mapping cardinality.

\begin{table}[h]
\centering
\small
\begin{adjustbox}{max width=\columnwidth}
\begin{tabular}{lrl}
\toprule
\textbf{FrenchNews-7 Class} & \textbf{IPTC Code} & \textbf{IPTC Media Topic} \\
\midrule
Culture \& Loisirs & 01000000 & Arts, Culture, Entertainment \& Media \\
(2 codes) & 10000000 & Lifestyle \& Leisure \\
\midrule
Économie (1 code) & 04000000 & Economy, Business \& Finance \\
\midrule
International & --- & \textit{no IPTC equivalent} \\
(0 codes) \\
\midrule
Politique (1 code) & 11000000 & Politics \& Government \\
\midrule
Sciences \& Technologies (1 code) & 13000000 & Science \& Technology \\
\midrule
Société & 14000000 & Society \\
(4 codes) & 02000000 & Crime, Law \& Justice \\
& 05000000 & Education \\
& 07000000 & Health \\
\midrule
Sport (1 code) & 15000000 & Sport \\
\bottomrule
\end{tabular}
\end{adjustbox}
\caption{Crosswalk between FrenchNews-7 classes and top-level IPTC Media Topics \citep{iptc_mediatopics_2025}. Per-class mapping cardinality shown in parentheses. 10 of 17 IPTC Level-1 categories map to a FrenchNews-7 class (58.8\% structural overlap). \textit{International} has no IPTC equivalent: it captures geographic scope (foreign vs.\ domestic routing), a dimension absent from the IPTC hierarchy.}
\label{tab:iptc_mapping}
\end{table}

\section{Exploratory Cross-Taxonomy Evaluation}
\label{app:iptc_experiment}

To quantify the taxonomy gap empirically, we probe EMMediaTopic \citep{kuzman2025} (an XLM-RoBERTa-large model trained on 21,000 non-French articles across 17 IPTC categories) on the FrenchNews-7 test split without retraining. Top-1 IPTC predictions are collapsed to FrenchNews-7 classes via Table~\ref{tab:iptc_mapping}; \textit{International} is excluded as it has no IPTC equivalent. Table~\ref{tab:iptc_experiment} reports macro-F1 both with and without unmapped-prediction penalty.

\begin{table}[h]
\centering
\small
\begin{adjustbox}{max width=\columnwidth}
\begin{tabular}{lrrr}
\toprule
\textbf{FrenchNews-7 Class} & \textbf{Support} & \textbf{F1 (all)} & \textbf{F1 (mapped)} \\
\midrule
Culture \& Loisirs & 2742 & 0.842 & 0.892 \\
Politique & 1825 & 0.771 & 0.796 \\
Sport & 1572 & 0.921 & 0.945 \\
Économie & 1311 & 0.635 & 0.696 \\
Société & 2034 & 0.591 & 0.736 \\
Sciences \& Technologies & 877 & 0.572 & 0.625 \\
\midrule
Macro-F1 (6 classes) & \textbf{--} & \textbf{0.722} & \textbf{0.781} \\
\bottomrule
\end{tabular}
\end{adjustbox}
\caption{EMMediaTopic classifier probed on FrenchNews-7 test split (13,146 articles, inference-only, no retraining). ``F1 (all)''\ treats unmapped IPTC predictions as incorrect; ``F1 (mapped)''\ evaluates only articles where the top-1 prediction maps to a FrenchNews-7 class. International excluded throughout.}
\label{tab:iptc_experiment}
\end{table}

\section{Reliability Tiers for Downstream Use}
\label{app:reliability_tiers}

Table~\ref{tab:reliability_tiers} maps three reliability tiers to concrete task types based on the cross-publisher evaluation results (\S\ref{sec:crosspub_gen}).

\begin{table*}[t]
\centering
\small
\setlength{\tabcolsep}{5pt}
\begin{adjustbox}{max width=\textwidth}
\begin{tabular}{lp{7.5cm}p{5.5cm}}
\toprule
Tier & Example Downstream Tasks & Basis \\
\midrule
\textbf{Safe} &
  Broad agenda-setting trend analysis; multi-outlet Sport / Culture \& Loisirs / International volume tracking; publisher editorial-specialization studies not dependent on economic or social-affairs distinctions &
  Cross-publisher recall $\geq 0.82$ for Sport, Culture \& Loisirs, International, Sciences \& Technologies \\
\addlinespace
\textbf{Use with caution} &
  Politique coverage comparisons across outlets; multi-class topic-distribution fingerprinting using all seven categories; longitudinal section-mix studies that treat Politique separately from Soci\'et\'e &
  Politique generalizes well but shares a soft boundary with Soci\'et\'e; cross-publisher macro-F1 $\approx 0.80$ \\
\addlinespace
\textbf{Requires supplementary validation} &
  Election-period economic framing studies; domestic-affairs agenda analysis where Soci\'et\'e volume is a primary signal; any study isolating \'Economie from Politique or Soci\'et\'e; longitudinal topic-count outcomes for \'Economie or Soci\'et\'e &
  \'Economie cross-publisher recall~=~0.517; Soci\'et\'e cross-publisher precision~=~0.577 \\
\bottomrule
\end{tabular}
\end{adjustbox}
\caption{Reliability-tier guidance for downstream use of the FrenchNews-7 classifier. Figures are from the held-out four-outlet cross-publisher evaluation (\S\ref{sec:crosspub_gen}).}
\label{tab:reliability_tiers}
\end{table*}

\section{Full-Context CamemBERTav2 Evaluation}
\label{app:full_context}

CamemBERTav2 supports a native 1,024-token context window but was originally evaluated under 512-token truncation for controlled comparison with CamemBERT-base. We retrained CamemBERTav2 at its full 1,024-token context (same hyperparameters: 6 epochs, lr~$= 2\times10^{-5}$, effective batch 32, seed~$= 42$) and evaluated on both the in-distribution test split and the held-out cross-publisher pool.

Table~\ref{tab:full_context_comparison} summarizes the comparison. At full context, CamemBERTav2 matches CamemBERT-base on both evaluation settings, confirming that the 0.4pp in-distribution gap reported under truncation in Table~\ref{tab:main_results} is an artifact of context-window limitation rather than architectural difference. CamemBERT-base remains the recommended deployment model.

\begin{table*}[t]
\centering
\small
\setlength{\tabcolsep}{5pt}
\renewcommand{\arraystretch}{1.12}
\begin{tabular*}{\textwidth}{@{\extracolsep{\fill}}lrrrr}
\toprule
\textbf{Model} & \textbf{Context} & \textbf{In-dist.\ Acc} & \textbf{In-dist.\ F1} & \textbf{Cross-pub F1} \\
\midrule
CamemBERT-base & 512 tok & 0.860 & 0.847 & 0.799 \\
CamemBERTav2 & 512 tok & 0.856 & 0.843 & --- \\
CamemBERTav2 & 1,024 tok & 0.861 & 0.847 & 0.798 \\
\bottomrule
\end{tabular*}
\caption{Comparison of CamemBERT-base (512 tokens) with CamemBERTav2 at 512-token truncation and its native 1,024-token context. In-distribution metrics on the 13,146-article test split; cross-publisher macro-F1 on the 2,100-article held-out outlet pool.}
\label{tab:full_context_comparison}
\end{table*}

\section{Pairwise Significance Tests}
\label{app:pairwise_significance}

Tables~\ref{tab:pairwise_indist} and~\ref{tab:pairwise_held_out} report paired bootstrap tests on macro-F1 and McNemar's tests on per-article correctness for all model pairs, with Holm--Bonferroni correction applied independently within each evaluation family. Bootstrap used 10,000 replicates with fixed seed~42. For McNemar's test, the continuity-corrected statistic $(|b-c|-1)^2/(b+c)$ is reported, where $b$ and $c$ are the off-diagonal discordant pair counts. Macro-F1 is averaged over the seven gold label classes; OOV predictions (one Llama-3.3-70B parse failure) are treated as incorrect without contributing a phantom class to the average.

\begin{table*}[t]
\centering
\footnotesize
\setlength{\tabcolsep}{4pt}
\renewcommand{\arraystretch}{1.05}
\begin{tabular*}{\textwidth}{@{\extracolsep{\fill}}llrrrcrrc}
\toprule
Model A & Model B & $\Delta$ F1 & \multicolumn{2}{c}{Bootstrap} & & \multicolumn{2}{c}{McNemar} & \\
\cmidrule(lr){4-5}\cmidrule(lr){7-8}
 & & & $p$ (Holm) & 95\% CI & Sig. & $\chi^2$ & $p$ (Holm) & Sig. \\
\midrule
CamemBERT-v1 & CamemBERTav2 & $+$0.004 & 0.086 & [$-$0.002,\ 0.009] &  & 2.11 & 0.147 &  \\
CamemBERT-v1 & mBERT & $+$0.014 & $<$0.001 & [0.008,\ 0.020] & $*$ & 31.26 & $<$0.001 & $*$ \\
CamemBERT-v1 & TF-IDF & $+$0.035 & $<$0.001 & [0.029,\ 0.042] & $*$ & 152.34 & $<$0.001 & $*$ \\
CamemBERTav2 & mBERT & $+$0.010 & 0.001 & [0.005,\ 0.016] & $*$ & 18.91 & $<$0.001 & $*$ \\
CamemBERTav2 & TF-IDF & $+$0.032 & $<$0.001 & [0.025,\ 0.038] & $*$ & 128.90 & $<$0.001 & $*$ \\
mBERT & TF-IDF & $+$0.021 & $<$0.001 & [0.015,\ 0.028] & $*$ & 56.70 & $<$0.001 & $*$ \\
\bottomrule
\end{tabular*}
\caption{In-distribution pairwise significance (13,146 test set, seed~42). Paired bootstrap (10,000 replicates) on macro-F1; McNemar's test on per-article correctness. Holm correction over all 12 raw p-values. $\Delta$ = macro-F1(A) $-$ macro-F1(B). $^*p < 0.05$ after correction.}
\label{tab:pairwise_indist}
\end{table*}

\begin{table*}[t]
\centering
\footnotesize
\setlength{\tabcolsep}{4pt}
\renewcommand{\arraystretch}{1.05}
\begin{adjustbox}{max width=\textwidth}
\begin{tabular}{llrrrcrrc}
\toprule
Model A & Model B & $\Delta$ F1 & \multicolumn{2}{c}{Bootstrap} & & \multicolumn{2}{c}{McNemar} & \\
\cmidrule(lr){4-5}\cmidrule(lr){7-8}
 & & & $p$ (Holm) & 95\% CI & Sig. & $\chi^2$ & $p$ (Holm) & Sig. \\
\midrule
CamemBERT & CamemBERTav2@1024 & $+$0.001 & 0.629 & [$-$0.012,\ 0.013] &  & 0.01 & 1.000 &  \\
CamemBERT & GPT-OSS-s11 & $+$0.043 & $<$0.001 & [0.027,\ 0.059] & $*$ & 31.49 & $<$0.001 & $*$ \\
CamemBERT & GPT-OSS-s22 & $+$0.043 & $<$0.001 & [0.027,\ 0.058] & $*$ & 32.56 & $<$0.001 & $*$ \\
CamemBERT & GPT-OSS-s33 & $+$0.039 & $<$0.001 & [0.023,\ 0.054] & $*$ & 25.73 & $<$0.001 & $*$ \\
CamemBERT & Mistral-Small-3.2 & $+$0.024 & 0.003 & [0.009,\ 0.040] & $*$ & 10.30 & 0.009 & $*$ \\
CamemBERT & Llama-3.3-70B & $+$0.028 & 0.002 & [0.012,\ 0.044] & $*$ & 6.56 & 0.035 & $*$ \\
CamemBERTav2@1024 & GPT-OSS-s11 & $+$0.042 & $<$0.001 & [0.027,\ 0.058] & $*$ & 30.88 & $<$0.001 & $*$ \\
CamemBERTav2@1024 & Mistral-Small-3.2 & $+$0.023 & 0.005 & [0.008,\ 0.039] & $*$ & 10.09 & 0.009 & $*$ \\
CamemBERTav2@1024 & Llama-3.3-70B & $+$0.027 & 0.001 & [0.012,\ 0.043] & $*$ & 6.85 & 0.035 & $*$ \\
GPT-OSS-s11 & Mistral-Small-3.2 & $-$0.019 & 0.001 & [$-$0.029,\ $-$0.009] & $*$ & 14.67 & 0.001 & $*$ \\
GPT-OSS-s11 & Llama-3.3-70B & $-$0.015 & 0.081 & [$-$0.031,\ 0.000] &  & 10.09 & 0.009 & $*$ \\
Mistral-Small-3.2 & Llama-3.3-70B & $+$0.004 & 0.629 & [$-$0.011,\ 0.018] &  & 0.37 & 1.000 &  \\
\bottomrule
\end{tabular}
\end{adjustbox}
\caption{Held-out pool pairwise significance (2,100 unseen-outlet pool, seed~42). GPT-OSS tested per seed against CamemBERT; seed~11 as representative for cross-model pairs. Holm correction within this family. $\Delta$ = macro-F1(A) $-$ macro-F1(B). $^*p < 0.05$ after correction. GPT-OSS-s11 vs.\ Llama: bootstrap CI includes 0 while McNemar $p = 0.009$; this disagreement reflects McNemar's sensitivity to correlated per-article errors that bootstrap macro-F1 averaging smooths over, but the direction (CamemBERT $>$ Llama) is consistent across all three GPT-OSS seeds.}
\label{tab:pairwise_held_out}
\end{table*}

\section{Case Study: Validating Editorial Fingerprints at Scale}
\label{app:case_study}

To demonstrate the utility of FrenchNews-7 as research infrastructure, we applied the fine-tuned CamemBERT-base model to a large-scale, unannotated longitudinal corpus (2022--2025). All topic distributions and divergence trends reported below are derived from classifier predictions, not gold labels; the in-distribution held-out test set (macro-F1 = 0.847 [0.841, 0.854], $n = 13{,}146$, same 13 outlets) provides the best available bound on prediction accuracy for this corpus.

\begin{figure}[h]
\centering
\includegraphics[width=\columnwidth]{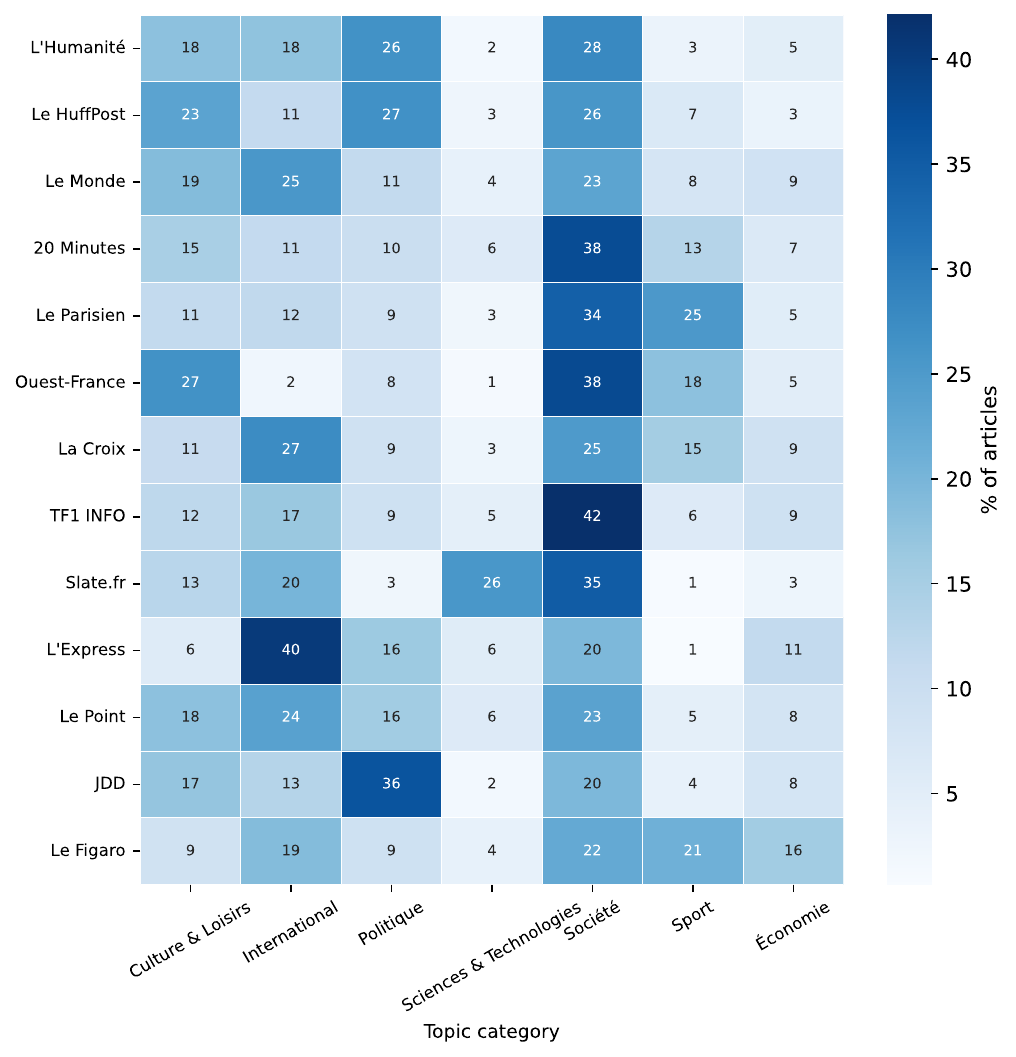}
\caption{Editorial topic fingerprints (2022--2025). Predicted topic shares for 13 outlets across 7 categories, aggregated from the unlabelled longitudinal corpus. Each cell shows the percentage of articles assigned to a given topic; rows sum to 100\%.}
\label{fig:topic_heatmap}
\end{figure}

Aggregating the predicted topic shares reveals distinct ``editorial fingerprints'' that align with known outlet identities, providing external validity evidence for our bottom-up taxonomy (see Figure~\ref{fig:topic_heatmap}). All 87,637 FrenchNews-7 manifest articles were excluded prior to aggregation by exact URL match; all statistics below are computed on the remaining 1,940,734 held-out articles only. \emph{Soci\'{e}t\'{e}} acts as a universal structural base (ranking first or second in all 13 outlets), while the maximum deviations reveal clear editorial specialisation. For example, the political weekly \emph{JDD} dedicates 36.4\% of its coverage to \emph{Politique} (over 10 points above the cross-outlet mean). Similarly, predicted distributions are consistent with the known international focus of \emph{L'Express} (40.5\% \emph{International}), the tech-and-ideas niche of \emph{Slate.fr} (25.7\% \emph{Sciences \& Technologies}), and the mass-market appeal of \emph{Le Parisien} (25.3\% \emph{Sport}). The regional paper \emph{Ouest-France} records near-zero predicted international coverage (2.3\%), consistent with its local remit.

\begin{figure}[h]
\centering
\includegraphics[width=\columnwidth]{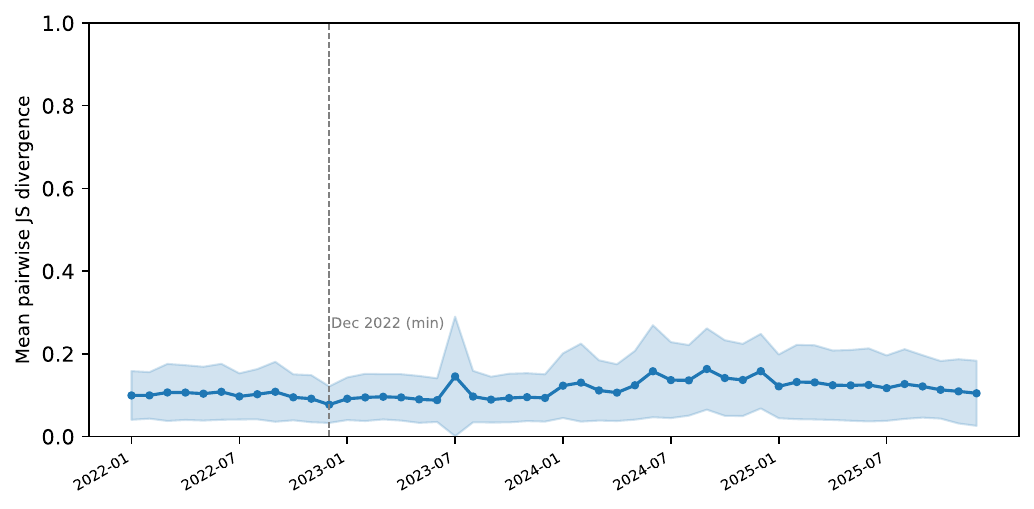}
\caption{Pairwise Jensen--Shannon divergence of outlet topic distributions over 48 months (2022--2025). Higher values indicate greater topical differentiation (see text for statistics). All values are computed from classifier predictions, not gold labels.}
\label{fig:js_divergence}
\end{figure}

Beyond static profiles, the classifier unlocks dynamic agenda-setting analysis \citep{mccombs1972,field2018}. By tracking the pairwise Jensen--Shannon (JS) divergence of topic distributions over 48 months, we observe a divergence trend: mean pairwise JS divergence increased from 0.098 [95\% CI: 0.094, 0.104] in the first 24 months (2022--2023) to 0.128 [0.122, 0.134] in the final 24 months (2024--2025), consistent with growing topical differentiation over this period (see Figure~\ref{fig:js_divergence}). CIs are bootstrap estimates (5,000 resamples of months within each window). Seasonal variation is visible within this trend: December 2022 records the lowest single-month divergence (0.077), potentially reflecting shared year-end and FIFA World Cup coverage that temporarily homogenised outlet agendas. This window also spans several major events that may independently shift outlet differentiation: the Russia--Ukraine war (from February 2022), the French presidential and legislative elections (2022), the Paris Olympics (July--August 2024), and the post-COVID news cycle recovery (2022--2023); disentangling genuine editorial drift from event-driven realignment would require a controlled design beyond the scope of this demonstration. These prediction-based trends are consistent with FrenchNews-7 capturing stable editorial boundaries, and illustrate the classifier's utility for longitudinal computational media research.

As a complementary temporal-generalization check, we evaluated the classifier against slug-derived reference labels on a stratified 100{,}000-article longitudinal sample from the same 13 training outlets (2022--2025). This check complements, rather than duplicates, the held-out four-outlet experiment in \S\ref{sec:crosspub_gen}: the held-out pool tests publisher transfer to unseen outlets, whereas the longitudinal slug check tests temporal stability on unseen later articles from the same 13 training outlets. Among the 63{,}245 slug-parseable articles (63.2\%), classifier accuracy is 82.1\% (95\% CI [81.8\%, 82.4\%]). Per-class recall ranges from 57.2\% for \emph{\'Economie} to 96.9\% for \emph{Sport}, and six of seven classes have Wilson half-widths at or below 1.3 percentage points; only \emph{Sciences \& Technologies} remains wider ($\pm 2.7$ points) because it is comparatively rare in the parseable subset (Table~\ref{tab:longitudinal_slug_check}). The $\sim$23-point gap between this recall (0.598) and the held-out pool result (0.833) reflects the balanced construction of the latter: the 300-article \emph{Sciences \& Technologies} slice in the held-out pool deliberately oversamples from outlets with structurally clean slugs, while the temporal corpus reflects the class's natural low prevalence and noisier slug coverage.

We then use this larger-sample accuracy estimate to calibrate the Monte Carlo error assumption on the target longitudinal corpus. The implied empirical error rate on the parseable subset is 17.9\%, and errors are again disproportionately absorbed by \emph{Soci\'{e}t\'{e}}. Non-parseable articles are enriched in \emph{Soci\'{e}t\'{e}} (+13.8 points) and \emph{\'Economie} (+5.5) and depleted in \emph{Sport} ($-13.0$) and \emph{International} ($-8.8$), so we retain $\epsilon = 0.20$ as the rounded central estimate and also report a stronger stress test at $\epsilon = 0.25$. Across 10{,}000 Monte Carlo iterations, the divergence increase from 2022--2023 to 2024--2025 remained strictly positive at both $\epsilon = 0.20$ (mean $\Delta = +0.0181$, 95\% CI [$+0.0172$, $+0.0189$]) and $\epsilon = 0.25$ (mean $\Delta = +0.0156$, 95\% CI [$+0.0147$, $+0.0164$]); the unperturbed estimate is $\Delta = +0.0296$. Increased noise attenuates the effect but does not approach sign reversal.

\begin{table}[h]
\centering
\small
\setlength{\tabcolsep}{5pt}
\renewcommand{\arraystretch}{1.05}
\begin{adjustbox}{max width=\columnwidth}
\begin{tabular}{lrrr}
\toprule
Category & $n$ & Recall & 95\% Wilson CI \\
\midrule
Culture \& Loisirs & 9,139 & 0.823 & [0.815, 0.831] \\
International & 11,326 & 0.868 & [0.861, 0.874] \\
Politique & 6,662 & 0.885 & [0.877, 0.893] \\
Sciences \& Technologies & 1,245 & 0.598 & [0.571, 0.625] \\
Société & 16,372 & 0.746 & [0.739, 0.752] \\
Sport & 12,965 & 0.969 & [0.966, 0.972] \\
Économie & 5,536 & 0.572 & [0.559, 0.585] \\
\bottomrule
\end{tabular}
\end{adjustbox}
\caption{Temporal-generalization check on the longitudinal corpus: per-class recall against slug-derived reference labels for the 63,245 parseable articles in a stratified 100,000-article sample from the same 13 training outlets (2022--2025). This evaluates later unseen articles from known publishers; it is complementary to, not a substitute for, the four-outlet held-out publisher test in \S\ref{sec:crosspub_gen}.}
\label{tab:longitudinal_slug_check}
\end{table}

\end{document}